\documentclass[journal,twoside,web]{ieeecolor}
\usepackage{tmi}
\usepackage{cite}
\usepackage{amsmath,amssymb,amsfonts}
\usepackage{algorithmic}
\usepackage{tabularx}
\usepackage{graphicx}
\usepackage{textcomp}
\usepackage{multirow}
\usepackage{booktabs}
\usepackage{bm}
\usepackage{stfloats}
\usepackage{array}
\usepackage{calrsfs}
\DeclareMathAlphabet{\mathcal}{OMS}{zplm}{m}{n}
\makeatletter
\let\NAT@parse\undefined
\makeatother
\usepackage[colorlinks=true,citecolor=blue,urlcolor=black]{hyperref}
\definecolor{myblue}{rgb}{0.2588, 0.5412, 0.8353}
\providecommand{\bluec}[1]{\textcolor{myblue}{#1}}
\definecolor{boxblue}{rgb}{0.5686, 0.6745, 0.8784}
\providecommand{\boxbluec}[1]{\textcolor{boxblue}{#1}}
\definecolor{segblue}{rgb}{0.6275, 0.6784, 0.8157}
\providecommand{\segbluec}[1]{\textcolor{segblue}{#1}}
\definecolor{segred}{rgb}{0.6510, 0, 0}
\providecommand{\segredc}[1]{\textcolor{segred}{#1}}
\definecolor{seggreen}{rgb}{0.1412, 0.4784, 0.4392}
\providecommand{\seggreenc}[1]{\textcolor{seggreen}{#1}}

\def\BibTeX{{\rm B\kern-.05em{\sc i\kern-.025em b}\kern-.08em
    T\kern-.1667em\lower.7ex\hbox{E}\kern-.125emX}}
\begin{document}
\title{ColoDiff: Integrating Dynamic Consistency\\With Content Awareness for\\Colonoscopy Video Generation}
\author{Junhu~Fu,
        Shuyu~Liang,
        Wutong~Li,
        Chen~Ma,
        Peng~Huang,
        Kehao~Wang,
        Ke~Chen,
        Shengli~Lin,
        Pinghong~Zhou,
        Zeju~Li,
        Yuanyuan~Wang, \textit{Senior Member, IEEE},
        and~Yi~Guo, \textit{Member, IEEE}
\thanks{This work was supported in part by the National Natural Science Foundation of China under Grant 62371139; and in part by the Shanghai Municipal Education Commission under Grant 24KNZNA09. \textit{(Corresponding authors: Zeju Li; Yuanyuan Wang; Yi Guo.)}}
\thanks{This work involved human subjects or animals in its research. Approval of all ethical and experimental procedures and protocols was granted by the Ethics Committee of Fudan University Shanghai Cancer Center, Shanghai, China, under Application No. 2509-Exp283, in 2025.}
\thanks{Junhu Fu, Shuyu Liang, Wutong Li, Chen Ma, Peng Huang, Kehao Wang, Zeju Li, Yuanyuan Wang, and Yi Guo are with the College of Biomedical Engineering, Fudan University, Shanghai 200433, China (e-mail: jhfu21@m.fudan.edu.cn; syliang22@m.\allowbreak fudan.edu.cn; wtli22@m.fudan.edu.cn; cma24@m.fudan.edu.cn; phuang22@m.\allowbreak fudan.\allowbreak edu.cn; wang\_kehao@fudan.\allowbreak edu.cn; zejuli@fudan.edu.cn; yywang@fudan.\allowbreak edu.cn; guoyi@fudan.edu.cn).}
\thanks{Ke Chen is with the Department of Endoscopy, Fudan University Shanghai Cancer Center, Shanghai 200032, China (e-mail: kechen23@m.fudan.edu.cn).}
\thanks{Shengli Lin and Pinghong Zhou are with the Endoscopy Center and Endoscopy Research Institute, Zhongshan Hospital, Fudan University, Shanghai 200032, China, and also with the Shanghai Collaborative Innovation Center of Endoscopy, Shanghai 200032, China (e-mail: lin.shengli@zs-hospital.sh.cn; zhou.pinghong@zs-hospital.sh.cn).}
}

\maketitle

\begin{abstract}
Colonoscopy video generation delivers dynamic, information-rich data critical for diagnosing intestinal diseases, particularly in data-scarce scenarios. High-quality video generation demands temporal consistency and precise control over clinical attributes, but faces challenges from irregular intestinal structures, diverse disease representations, and various imaging modalities. To this end, we propose ColoDiff, a diffusion-based framework that generates dynamic-consistent and content-aware colonoscopy videos, aiming to alleviate data shortage and assist clinical analysis. At the inter-frame level, our TimeStream module decouples temporal dependency from video sequences through a cross-frame tokenization mechanism, enabling intricate dynamic modeling despite irregular intestinal structures. At the intra-frame level, our Content-Aware module incorporates noise-injected embeddings and learnable prototypes to realize precise control over clinical attributes, breaking through the coarse guidance of diffusion models. Additionally, ColoDiff employs a non-Markovian sampling strategy that cuts steps by over 90\% for real-time generation. ColoDiff is evaluated across three public datasets and one hospital database, based on both generation metrics and downstream tasks including disease diagnosis, modality discrimination, bowel preparation scoring, and lesion segmentation. Extensive experiments show ColoDiff generates videos with smooth transitions and rich dynamics. ColoDiff also produces customized contents tailored for diverse tasks, e.g., colitis, polyps, and adenomas for diagnosis. Incorporating synthetic videos into training promotes discriminative representation learning and improves diagnosis accuracy by 7.1\%. ColoDiff presents an effort in controllable colonoscopy video generation, revealing the potential of synthetic videos in complementing authentic representation and mitigating data scarcity in clinical settings.
\end{abstract}

\begin{IEEEkeywords}
Colonoscopy video generation, diffusion model, temporal consistency, content controllability.
\end{IEEEkeywords}

\section{Introduction}
\label{sec: intro}
\definecolor{amber(sae/ece)}{rgb}{1.0, 0.49, 0.0}
\begin{figure*}[!htbp]
\centering
\includegraphics[width=1.0\textwidth]{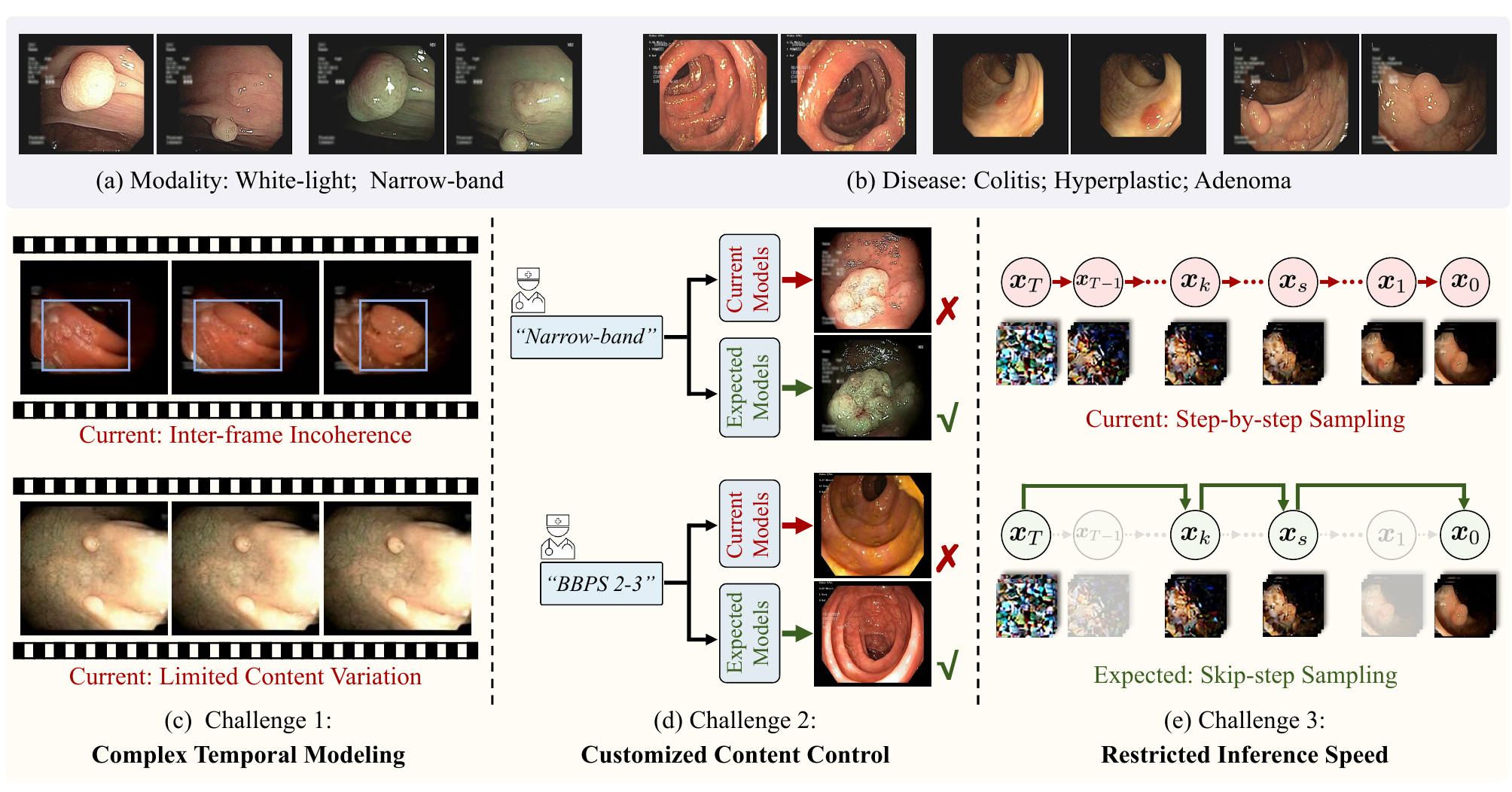}
\caption{\bluec{(a)}-\bluec{(b)} Colonoscopy video analysis serves as an important diagnostic approach, integrating diverse imaging modalities and disease manifestations. \bluec{(c)}-\bluec{(e)} Existing challenges for colonoscopy video generation.}
\label{Fig1_challenge}
\end{figure*}

\IEEEPARstart{C}{olorectal} diseases often present with subtle early symptoms yet remain a leading cause of global mortality~\cite{CRC}, thus demanding early intervention and improved prognosis. Colonoscopy video analysis plays a crucial role in gastrointestinal diagnosis, providing dynamic, multi-perspective views into mucosal microvascular structures and offering real-time feedback. These capabilities form a solid foundation for a wide range of clinical tasks, from bowel preparation scoring~\cite{hyperkvasir,qualitycontrol}, lesion screening or tracking~\cite{SUN-SEG}, to disease diagnosis~\cite{IPNet}, as well as auxiliary tasks like modality discrimination~\cite{modalitydis2,modalitydis1}. Example colonoscopy video sequences with different modalities or diseases are illustrated in Fig.~\ref{Fig1_challenge}(a)-(b), further demonstrating the strength of video analysis in capturing temporal-coherent and multi-view features for diverse intestinal conditions. Deep learning methods trained on large-scale data have achieved remarkable success in visual analysis, but collecting adequate colonoscopy videos is impractical in clinical scenarios due to privacy regulation, laborious annotation, and heterogeneous protocols. The lack of high-quality data severely constrains computer-aided diagnosis and treatment, underscoring the urgent need for effective solutions.

Colonoscopy data generation has emerged as a promising solution to bridge this gap. In the broader field of image generation, diffusion-based approaches dominate with superior synthesis quality and training stability~\cite{diffusion,DDPM}. As for colonoscopy image generation, recent algorithms~\cite{ControlPolyp,coloimage1,coloimage2,coloimage3} build on latent diffusion models (LDMs)~\cite{LDM}, leveraging variational autoencoder (VAE) latent spaces~\cite{Taming,VAE} for perceptual compression and high-fidelity synthesis. Despite these advances, the models focus on 2D static generation, failing to capture multi-perspective views and dynamic information. In the realm of video generation, transition from static to dynamic requires modeling temporal dependencies for inter-frame coherence. Some approaches employ 3D U-Net to jointly model spatio-temporal features~\cite{VDM,LVDM}, and others propose continuous temporal encoding with motion disentanglement~\cite{GAN,styleganv,mostganv}. For colonoscopy video generation specially, it faces more intricate dynamics from irregular intestinal structures, coupled with variable mucosal and microvascular manifestations across imaging modalities and diseases. Although the latest diffusion-based methods show promise in endoscopic video generation by providing synthetic unlabeled data for downstream tasks, they still lack inter-frame consistency and content controllability~\cite{Endora,FEAT}, underscoring the need for customized representation and targeted generation.

While recent efforts demonstrate the feasibility of colonoscopy video generation, there remain some critical challenges: (1) \textbf{Complex temporal modeling.} Existing methods employ 3D operations~\cite{VDM,LVDM} or concatenate frames into pseudo-large volumes~\cite{ControlVideo} for feature extraction, but both inadequately capture temporal dynamics, leading to inter-frame inconsistency for intricate scenarios like irregular morphology and scale-varying anatomy (Fig.~\ref{Fig1_challenge}(c)). (2) \textbf{Customized content control.} Conditional diffusion models rely on the time-step index for noise perception and fixed encodings for diverse representations~\cite{diffusion,DDPM}. Such information is insufficient for colonoscopy videos~\cite{hyperkvasir,SUN-SEG,colonoscopic}, which involve various disease manifestations and imaging modalities, thereby hindering controllable generation (Fig.~\ref{Fig1_challenge}(d)). (3) \textbf{Restricted inference speed.} Diffusion-based video generation requires extensive temporal modeling and hundreds of sampling steps, thus preventing real-time inference (Fig.~\ref{Fig1_challenge}(e)).

In this paper, we propose ColoDiff, a diffusion-based framework for dynamic-consistent and content-aware colonoscopy video generation. ColoDiff integrates tailored temporal modeling, precise content control, and an efficient non-Markovian sampling strategy, enabling real-time synthesis while addressing the limitations of existing methods. In summary, the main contributions of this paper are as follows:
\begin{enumerate}
	\item[(i)] We propose ColoDiff, a diffusion-driven architecture that integrates TimeStream and Content-Aware modules for real-time colonoscopy video generation. By addressing the challenges of temporal modeling, content controllability, and inference efficiency, ColoDiff establishes a new framework for supplementing specific representation and mitigating data scarcity.

    \item[(ii)] At the inter-frame level, our TimeStream module explicitly decouples temporal dependencies from video sequences through a cross-frame tokenization mechanism, capturing motion patterns during endoscope movement. At the intra-frame level, our Content-Aware module incorporates noise-injected embedding and category-aware prototypes for fine-grained modulation, enabling precise control over clinical attributes.

    \item[(iii)] Extensive experiments of generated videos demonstrate enhanced temporal consistency and controllability, with synthetic-augmented data improving disease diagnosis accuracy by 7.1\% and segmentation Dice by 6.2\%. The results prove ColoDiff’s groundbreaking contribution in complementing customized videos and improving downstream task performance.
\end{enumerate}

The above is the Introduction section of this paper. The Related Works, Methodology, Experiments and Results, Discussion, and Conclusion sections are followed in turn.
\section{Related Works}
\label{sec: related_works}
\subsection{From Static to Dynamic}
Medical image generation, including colonoscopy image synthesis, has become a privacy-preserving approach to mitigate data scarcity. Recent algorithms~\cite{ControlPolyp,coloimage1,coloimage2,coloimage3} combine the generative power of diffusion process with the efficiency of perceptual compression~\cite{Taming,VAE}, enabling the synthesis of high-fidelity images for improved diagnosis. Despite the advances, these methods remain confined to 2D static generation, lacking the ability to capture multi-perspective views and temporal dynamics. This limitation has motivated a growing shift toward dynamic generation, where temporal information is explicitly modeled to better reflect clinical scenarios.

Following the trend of dynamic modeling, some algorithms attempt to extend static image models to video synthesis by incorporating temporal constraints~\cite{tunevideo,makevideo,alignlatents}. These approaches typically integrate fine-tuned temporal layers into existing image models, reducing the need for large-scale video datasets. At the same time, this pattern is inherently limited by the static prior of pre-trained image models, struggling to capture dynamic characteristics, particularly complex motion patterns and long-term temporal dependencies. The key-frame-first approach with intermediate frame interpolation~\cite{zuointerpolation,AAAIinterpolation}, has also been proposed to enhance temporal coherence, while this strategy may lead to inconsistent motion trajectories and fail to introduce effective information. Other algorithms suggest training video generation models from scratch, which allows direct learning of temporal dynamics without static bias. In this setting, U-Net and Transformer architectures remain the predominant choices. Ho et al. and He et al. utilize the standard diffusion setup with a 3D U-Net architecture for video generation, while facing inherent inductive bias like locality and translation equivariance~\cite{VDM,LVDM}. Additionally, the 3D convolution substantially inflates parameter scales. To achieve better scalability, Sora and Latte leverage a diffusion Transformer (DiT) architecture that operates on spacetime patches of video latent encoding~\cite{Latte,DiT}. Zhang et al. propose fully cross-frame interaction, which concatenates all video frames into a large image for joint encoding via Transformer, but this increases the number of input tokens, leading to high computational complexity~\cite{ControlVideo}.

As can be seen, while the transition from static to dynamic has become an important direction, medical video generation still struggles with effectively modeling temporal dynamics. How to achieve inter-frame consistency while maintaining manageable model scale and computational efficiency remains a problem to be solved.

\definecolor{amber(sae/ece)}{rgb}{1.0, 0.49, 0.0}
\begin{figure*}[!htbp]
\centering
\includegraphics[width=0.85\textwidth]{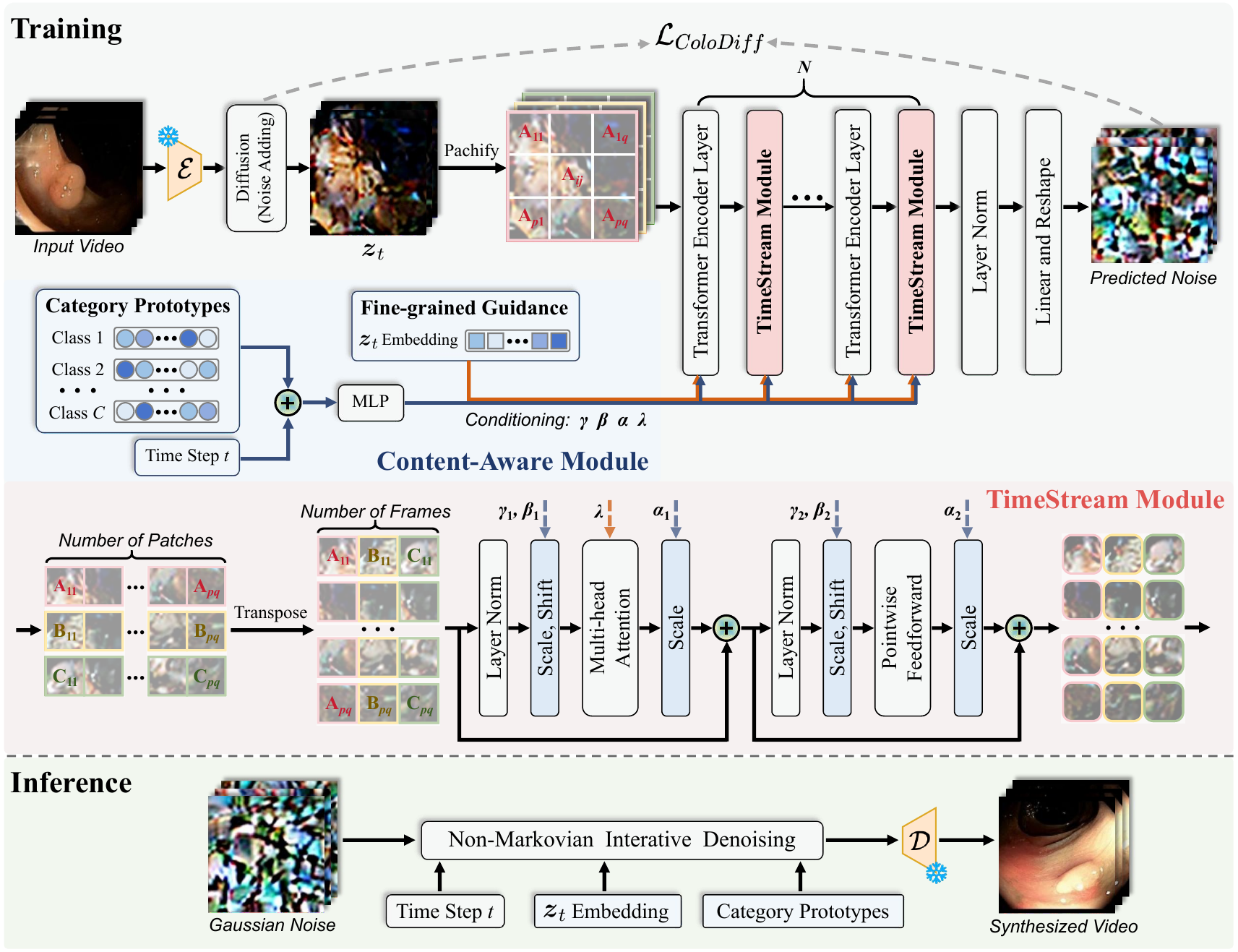}
\caption{The overall workflow of ColoDiff. $\mathcal{E}$ and $\mathcal{D}$ denote pre-trained encoder and decoder in VAE, respectively. $\boldsymbol{z}_t$ represents the latent features that have been added noise. The light red area shows how the TimeStream module decouples temporal information, and the light blue area shows how the Content-Aware module controls generation process.}
\label{Fig2_method}
\end{figure*}

\subsection{From Uncontrollable to Customized}
The latest diffusion-based methods have made promising attempts in endoscopic video generation~\cite{Endora,FEAT}. Due to the lack of textual descriptions and category labels, the synthetic videos remain uncontrollable. They cannot reliably reflect specific disease types, imaging modalities, or other clinically relevant attributes, and thus are used for unlabeled data augmentation in semi-supervised downstream tasks~\cite{Endora,FEAT}. This limitation highlights the demand for methods that support customized representation and targeted generation.

In the broader field of controllable natural video generation, current approaches primarily rely on textual and visual alignment. For diffusion models based on U-Net, Stable Diffusion injects condition information by inserting cross-attention modules into intermediate layers~\cite{LDM}, while ControlNet adds trainable condition branches to pre-trained U-Net structure~\cite{ControlNet}. For diffusion models based on Transformer, U-ViT encodes conditioning as extra tokens that are concatenated with visual tokens and processed through Transformer blocks with long skip connections~\cite{UViT}. Similarly, DiT employs a pure Transformer architecture that incorporates conditioning via adaptive layer normalization, progressively increasing conditional influence during training through zero-initialization strategies~\cite{DiT}. To sum up, conditioning in existing diffusion models typically includes time-step indices, textual embedding, and class embedding. 

Unlike natural data, medical videos are inherently constrained by limited data and lack of paired texts~\cite{Bora}, making it difficult to learn robust representations of diverse patterns and achieve high controllability for customized contents. In the absence of textual guidance, time-step and class embedding become the only remaining conditioning, but their coarse granularity precludes fine-grained alignment for precise control.

\section{Methodology}
\label{sec: method}

Fig.~\ref{Fig2_method} illustrates the workflow of ColoDiff, our diffusion-based model with Transformer architecture, supporting colonoscopy video generation with dynamic consistency and content controllability. During training, ColoDiff learns to predict added noise in the diffusion process under conditional guidance, which enables Transformer to learn the denoising pathway. Specifically, the TimeStream module decouples temporal dynamics via inter-frame interactions to guarantee video coherence, while the Content-Aware module leverages noise-injected embedding with trainable prototypes for intra-frame clinical attribute control. During inference, ColoDiff achieves customized generation through a non-Markovian chain, guiding the denoising process from Gaussian noise.

\subsection{TimeStream Module Enhances Dynamic Consistency}
Colonoscopy videos present complex spatio-temporal patterns due to irregular intestinal structures, which challenges temporal consistency modeling. Although Transformers~\cite{DiT,Transformer,ViT} excel at capturing long-range dependency within a frame, efficient cross-frame interaction for temporal modeling remains computationally expensive~\cite{ControlVideo}. To address this limitation, we introduce the TimeStream module, specifically designed to enhance temporal coherence by modeling inter-frame relationships in a computationally efficient manner.

As illustrated in Fig.~\ref{Fig2_method}, the Transformer encoder with $D$ latent dimensions would produce feature maps $\boldsymbol{z}_t^i \in \mathbb{R}^{F \times P \times D}$, where $i$ represents the layer index, $F$ represents the number of frames and $P$ denotes the number of patches per frame. Then we use a transpose-like operation to rearrange all tokens, considering patches with identical spatial location across different frames as sequential inputs. These patches tend to reflect the movement of the same anatomical structures, e.g., specific lesions or capillaries. Therefore, $P$ sequences with length $F$ are formed, and the shape of feature is transformed to $\mathbb{R}^{P \times F \times D}$. In this way, we can feed these $P$ sequences into the subsequent blocks, including multi-head attention (MHA) and MLP-based feedforward layers, in parallel:
\begin{equation}
     \boldsymbol{h}_t^i = \text{MHA} \left\{ \text{LayerNorm} \left[  \left( \boldsymbol{z}_t^{i} \right) ^T \right] \right\} + \left( \boldsymbol{z}_t^{i} \right) ^T
    \label{eq:mha},
\end{equation}
\begin{equation}
    \boldsymbol{z}_t^{i+1} = \text{MLP}\left[\text{LayerNorm}\left(\boldsymbol{h}_t^i\right)\right] + \boldsymbol{h}_t^i
    \label{eq:mlp_layernorm},
\end{equation}
where $\boldsymbol{h}_t^{i}$ denotes the hidden state at the $i$-th layer, and $\left( \boldsymbol{z}_t^{i} \right) ^T \in \mathbb{R}^{P \times F \times D}$ denotes the transpose of $\boldsymbol{z}_t^{i}$. Specifically, in each of the $P$ sequences, the $F$ tokens are from different frames. These tokens initially undergo an attention mechanism, aggregating information from all other tokens via learned attention weights to compute temporal context-aware representations. Following this, they pass through residual connection and layer normalization before entering an MLP that applies a position-wise transformation. The MLP uses a non-linear GELU activation, expanding feature dimensionality before projecting back to the original dimension, and enables each token to integrate both long-range and short-range inter-frame connections. The output $\boldsymbol{z}_t^{i+1}$ of the $i$-th layer is finally produced after another round of residual connection and normalization, ensuring stable gradients throughout the deep network while enabling complex interactions. To avoid spatial information loss, we employ Transformer encoder layers and TimeStream modules in sequence, with one followed by another, totally repeated $N$ ($N$=28) times (see top Fig.~\ref{Fig2_method}).

The TimeStream module leverages prior knowledge that the same anatomical structure generally occupies consistent or adjacent spatial locations across consecutive frames, and its variation over time is continuous and modelable. This enables accurate modeling of irregular intestinal structures and dynamic motion patterns. In addition, this module empowers ColoDiff to efficiently model temporal dynamics by using 2D models for 3D contextual reasoning, without increasing model scale or computational cost.

\subsection{Content-Aware Module Provides Precise Guidance}
To ensure clinical utility, generated colonoscopy videos should align with specific clinical attributes, which requires the content-aware ability of ColoDiff. As a general principle, diffusion models should adapt to inputs with varying noise levels across diverse time steps, but existing frameworks only rely on the time-step index $t$ to perceive noise levels, yielding coarse conditioning without content-aware information.

To address this limitation, our Content-Aware module introduces the input video embedding as additional guidance. Specifically, the noise-injected data $\boldsymbol{z}_t$ is encoded into $\text{Embed} \left( \boldsymbol{z}_t^i \right)$ by a learnable encoder. Compared with time-step index $t$ only providing a global representation for noise level, $\text{Embed} \left( \boldsymbol{z}_t^i \right)$ merges information from noise level, intra-frame visual concepts, and their interaction after noise injection. Hence, it serves as a fine-grained condition with intra-frame spatial information for ColoDiff’s Transformer blocks. This enriched conditioning revises the attention mechanism as
\begin{equation}
    \text{Softmax} \left[ \frac{\boldsymbol{Q}^i (\boldsymbol{K}^i)^T}{\sqrt{d_k}} \right] \left[ \boldsymbol{V}^i + \lambda \cdot \text{Embed} \left( \boldsymbol{z}_t^i \right) \right],
    \label{eq:attention}
\end{equation}
where $\boldsymbol{Q}^i$, $\boldsymbol{K}^i$, and $\boldsymbol{V}^i$ denote the query, key, and value at the $i$-th layer. $d_k$ denotes the dimension of query and key, serving as a scaling factor. $\lambda$ is a learnable weighting parameter with zero-initialization, ensuring that $\text{Embed} \left( \boldsymbol{z}_t^i \right)$ produces a progressive influence. According to (\ref{eq:attention}), ColoDiff can better perceive the injected noise level by integrating $\text{Embed} \left( \boldsymbol{z}_t^i \right)$ into the attention mechanism, thus achieving improved noise prediction and content-aware modulation.

In addition, to make ColoDiff obtain more specific representation from various video patterns, our Content-Aware module introduces prototype learning. Different from current fixed encoding schemes~\cite{DiT,Transformer}, we assign a learnable representation vector, i.e., prototype, for each category, as illustrated in Fig.~\ref{Fig2_method}. Based on these prototypes, ColoDiff employs scaling parameters $\gamma$, $\alpha$ and bias parameter $\beta$ to regulate affine transformations of multi-layer features, thus controlling the content of generated videos. These parameters are also zero-initialized, which ensures training stability while exerting incremental influence of condition information. (\ref{eq:norm}) and (\ref{eq:recover}) present the detailed forward data processing flow, where $f^n$ denotes the $n$-th feature of an input sample; the normalized feature value $\tilde{{f}^{n}}$ is scaled and shifted under the modulation of $\gamma$ and $\beta$ to produce the final feature $f_{\mathrm{out}}^{n}$:
\begin{equation}
    \tilde{{f}^{n}}=\frac{f^{n}-\mu}{\sqrt{\sigma^{2}+\epsilon}},
    \label{eq:norm}
\end{equation}
\begin{equation}
    f_{\mathrm{out}}^{n}=\gamma \cdot \tilde{{f}^{n}}+\beta.
    \label{eq:recover}
\end{equation}
The scaling mechanism of $\alpha$ is identical to that of $\gamma$. During training, as ColoDiff progressively fits the noise patterns specific to different categories, the prototypes become infused with class-discriminative information.

To summarize, the noise-injected video embedding enhances fine-grained visual perception beyond noise level, while the learnable prototypes offer more flexible and class-discriminative representations. By integrating them together, our Content-Aware module enables synthetic videos to accurately reflect controllable clinical attributes.

\subsection{ColoDiff With Non-Markovian Sampling}
We build ColoDiff on a state-of-the-art (SOTA) diffusion-based generative model~\cite{DDPM,DDIM}, which learns data distribution through a gradual diffusion and denoising process. The framework involves a forward process $q$ that incrementally corrupts data by adding Gaussian noise over multiple time steps and a reverse process $p_\theta$ that iteratively denoises samples to recover data similar to the original distribution, where $\theta$ represents learnable model parameters. Specifically, at a time step $t$ in the forward process, a noisier image $\boldsymbol{x}_{t}$ is calculated from a cleaner image $\boldsymbol{x}_{t-1}$ by adding noise:
\begin{equation}
    q(\boldsymbol{x}_t|\boldsymbol{x}_{t-1}) = \mathcal{N}(\boldsymbol{x}_t|\sqrt{1-\beta_t}\boldsymbol{x}_{t-1},\beta_t \boldsymbol{I}),
    \label{equ1}
\end{equation}
where $\mathcal{N}$ denotes the Gaussian distribution, $\boldsymbol{I}$ denotes the identity matrix, and $\beta_t \in (0,1)$ is a step-varying hyper-parameter. The reverse process uses a noise estimator, $\epsilon_{\theta}(\boldsymbol{x}_{k}, k)$, which predicts the noise within a noisy image to enable its reconstruction into a clean image.

Diffusion models typically rely on a Markovian reverse process that denoises data over hundreds of steps for high-quality generation. While acceptable for images, this is computationally prohibitive for videos. ColoDiff addresses this by implementing a non-Markovian reverse process~\cite{DDIM}. The model first uses its current state $\boldsymbol{x}_k$ and the time step $k$ to predict an estimate of the clean image $\hat{\boldsymbol{x}}_0$, which is a linear combination of the noisy input $\boldsymbol{x}_k$ and the predicted noise $\epsilon_{\theta}(\boldsymbol{x}_{k}, k)$. Crucially, with this estimate $\hat{\boldsymbol{x}}_0$ in hand, the reverse process can then reconstruct any previous state $\boldsymbol{x}_s$ (where $s < k$) in a single step. The integration of non-Markovian strategy allows the sampler to jump between non-adjacent time steps, dramatically accelerating inference~\cite{DDIM}. As a result, ColoDiff can generate high-quality videos in real-time by using a drastically reduced number of sampling steps.

Additionally, we choose to perform computation in the latent space~\cite{LDM}. It means that during the earlier process, the input video $\boldsymbol{x}_0$ is first encoded into a low-dimensional space as $\boldsymbol{z}_0$ by a pre-trained VAE~\cite{VAE} with encoder $\mathcal{E}$. Similarly, in the later process, the final inference will be decoded back to the image space with a deterministic pass through decoder $\mathcal{D}$. Thus, the final loss function of ColoDiff with conditioning regulation can be formulated as~\cite{loss1,DDPM,LDM}: 
\begin{equation}
    \mathcal{L}_{ColoDiff}=\mathbb{E}_{\boldsymbol{z}_{0}, \epsilon \sim \mathcal{N}(0,1), t}\left[\left\|\epsilon-\epsilon_{\theta}\left(\boldsymbol{z}_{t}, \boldsymbol{c}, t\right)\right\|_{2}^{2}\right],
    \label{eq:Loss}
\end{equation}
where $t$ is uniformly sampled from $\{1, \ldots, T\}$, $\boldsymbol{c}$ denotes the embedded control information from $\text{Embed}(\boldsymbol{z}_t^i)$ and learnable prototypes, $\epsilon$ represents the target noise, and $\epsilon_{\theta}$ is the model that learns noise.

The skip-step sampling enables ColoDiff to drastically reduce inference steps while maintaining generation quality, and the latent representation further improves computational efficiency. These designs make real-time colonoscopy video generation feasible in clinical settings.

\section{Experiments and Results}
\label{sec: experiments}
\subsection{Experimental Settings}
\subsubsection{Datasets}
We collect a total of 4,597 labeled colonoscopy video clips from three public datasets and one hospital database, each corresponding to different tasks. The Colonoscopic dataset contains 152 original videos, annotated with both disease types and imaging modalities, making it well-suited for disease classification and modality discrimination~\cite{colonoscopic}. The HyperKvasir dataset, with 373 original videos, provides annotations of disease types together with Boston Bowel Preparation Scale (BBPS) scores, enabling both disease classification and bowel preparation scoring~\cite{hyperkvasir}. Specifically, the disease labels cover colitis, polyp, and adenoma, while the modality labels distinguish between white-light and narrow-band imaging (WLI \textit{vs.} NBI). WLI provides a brighter view for clinicians during screening, while NBI displays clearer mucosal and microvascular structures during diagnosis~\cite{D2polyp}. The bowel preparation scores are categorized into BBPS 0-1 and BBPS 2-3, reflecting varying levels of intestinal cleanliness. The SUN-SEG dataset focuses on dense prediction, offering 285 original videos with frame-by-frame segmentation masks, resulting in 49,136 frames for pixel-level annotation~\cite{SUN-SEG}. In addition, we collect 578 original videos from Fudan University Shanghai Cancer Center. This hospital database was approved by the Ethics Committee of Fudan University Shanghai Cancer Center, Shanghai, China (No. 2509-Exp283) in 2025. The example videos have been displayed in Fig.~\ref{Fig1_challenge}(a)-(b).

For comparison with SOTA methods, experiments are conducted on each dataset to make a fair evaluation. For ablation and downstream tasks, experiments are conducted on specific tasks. Taking modality discrimination task as an example, the used WLI and NBI videos are all from Colonoscopic dataset. Regarding data allocation, Colonoscopic, HyperKvasir, and hospital datasets follow an 8:2 split, with 80\% videos randomly selected for training and the rest for evaluation. The SUN-SEG dataset is separated according to official guidelines.

\subsubsection{Implementation Details}
We conduct all experiments using the PyTorch framework and employ two NVIDIA A100 GPUs for training. The input videos are uniformly adjusted to 128$\times$128 with 16 frames. We use the pre-trained VAE based on LDM~\cite{LDM}, with a latent space of 16$\times$16$\times$4. During the training process, the upper limit of iterations is set to be 500,000 with an early stop mechanism to avoid overfitting. We choose a batch size of 32, a learning rate of 1$\times10^{-4}$, and the AdamW optimizer. Additionally, we implement an exponential moving average (EMA) strategy~\cite{UViT,DiT} for model parameters, ensuring training stability and output consistency. Prototype vectors match the input category count in number, with each vector dimension configured to 1,024. With these settings, ColoDiff converges in an average training time of 6.5 hours. Hyper-parameters for other compared algorithms follow the specifications in~\cite{LVDM,Endora,mostganv,styleganv,FEAT}.

\begin{table*}[t]
    \centering
    \caption{Quantitative Comparison With SOTA Algorithms. Best Results Are in \textbf{Bold}.}
    \begin{tabular}{c c *{12}{c}}
        \toprule
        \multirow{2}{*}{Type} & \multirow{2}{*}{Method} & 
        \multicolumn{3}{c}{Colonoscopic \cite{colonoscopic}} & 
        \multicolumn{3}{c}{HyperKvasir \cite{hyperkvasir}} & 
        \multicolumn{3}{c}{SUN-SEG \cite{SUN-SEG}} &
        \multicolumn{3}{c}{Hospital Data} \\
        \cmidrule(lr){3-5} \cmidrule(lr){6-8} \cmidrule(lr){9-11} \cmidrule(lr){12-14}
        & & FVD$\downarrow$ & FID$\downarrow$ & IS$\uparrow$ & 
        FVD$\downarrow$ & FID$\downarrow$ & IS$\uparrow$ & 
        FVD$\downarrow$ & FID$\downarrow$ & IS$\uparrow$ &
        FVD$\downarrow$ & FID$\downarrow$ & IS$\uparrow$ \\
        \midrule
        \multirow{2}{*}{GAN-based} 
        & StyleGAN-V \cite{styleganv} & 2111 & 226.1 & 2.12 & 729 & 88.5 & 1.98 & 682 & 46.5 & 3.91 & 703 & 51.7 & 3.42 \\
        & MoStGAN-V \cite{mostganv} & 469 & 53.2 & 3.37 & 607 & 29.4 & 2.74 & 412 & 28.6 & 4.04 & 522 & 46.6 & 3.59 \\
        \midrule
        \multirow{4}{*}{Diffusion-based} 
        & LVDM \cite{LVDM} & 1037 & 96.9 & 1.93 & 528 & 49.7 & 2.15 & 387 & 35.9 & 3.77 & 416 & 40.3 & 3.12 \\
        & Endora \cite{Endora} & 461 & 13.4 & 3.90 & 496 & 19.6 & 3.29 & 375 & 17.8 & 4.59 & 389 & 23.2 & 3.65 \\
        & FEAT-L \cite{FEAT} & 351 & \textbf{12.3} & \textbf{4.01} & 511 & 21.2 & 3.27 & 356 & 13.6 & 4.61 & 402 & 18.9 & 3.88 \\
        & ColoDiff (Ours) & \textbf{339} & 12.7 & 3.95 & \textbf{473} & \textbf{16.3} & \textbf{3.46} & \textbf{294} & \textbf{11.9} & \textbf{4.78} & \textbf{336} & \textbf{15.7} & \textbf{4.14} \\
        \bottomrule
    \end{tabular}
    \label{tab1:comparison}
    \vspace{2mm}
\end{table*}

\subsubsection{Evaluation Metrics}
For comparison and ablation experiments, we adopt metrics including Fréchet Inception Distance (FID)~\cite{FID}, Fréchet Video Distance (FVD)~\cite{FVD}, and Inception Score (IS)~\cite{IS}. To be specific, FID evaluates the fidelity of generated data, computing the Fréchet distance between two multi-variate Gaussian distributions fitted to deep features of real and generated images:
\begin{equation}
\text{FID} = \|\mu_r - \mu_g\|_2^2 + \text{Tr}(\Sigma_r + \Sigma_g - 2(\Sigma_r \Sigma_g)^{1/2}),
\label{eq:fid}
\end{equation}
where $(\mu_r, \Sigma_r)$ and $(\mu_g, \Sigma_g)$ are the mean and covariance of the real and generated image features, respectively. FVD extends this idea to the video domain. By employing the same scheme but calculating mean and covariance from spatio-temporal features, it evaluates the temporal consistency of generated videos. Additionally, IS goes beyond fidelity, measuring the diversity and controllability of generated data. It calculates the KL divergence between conditional class distributions and marginal class distributions:
\begin{equation}
    \text{IS} = \exp \left( \mathbb{E}_{x \sim p_g} \left[ D_{\text{KL}}(p(y|x) \parallel p(y)) \right] \right),
    \label{eq:iscore}
\end{equation}
where $p_g$ denotes the distribution of generated samples, $p(y|x)$ is the class distribution of the generated data $x$, and $p(y)$ is the marginal class distribution. Lower FID and FVD values indicate better fidelity and temporal consistency, while higher IS values reflect greater diversity and controllability.

For downstream tasks, we use precision, recall, F1-score, and accuracy for classification; as well as Dice, mIoU, precision, and recall for segmentation. Together, these metrics comprehensively assess both the generative quality of ColoDiff and its performance in supporting clinical downstream tasks.

\subsection{Comparison with SOTA Methods}
\begin{table*}[t]
    \centering
    \caption{Ablation Experiment Results for TimeStream And Content-Aware Modules. Best Results Are in \textbf{Bold}.}
    \begin{tabular}{c c *{9}{c}} 
        \toprule
        \multicolumn{2}{c}{Category} & 
        \multicolumn{3}{c}{Colitis} & 
        \multicolumn{3}{c}{Polyp} & 
        \multicolumn{3}{c}{Adenoma} \\
        \cmidrule(lr){3-5} \cmidrule(lr){6-8} \cmidrule(lr){9-11}
        \multicolumn{2}{c}{Metric} & 
        FVD$\downarrow$ & FID$\downarrow$ & IS$\uparrow$ & 
        FVD$\downarrow$ & FID$\downarrow$ & IS$\uparrow$ & 
        FVD$\downarrow$ & FID$\downarrow$ & IS$\uparrow$ \\
        \midrule
        
        \multirow{2}{*}{Temporal Coherence} 
        & Transformer Encoder Layer & 518 & 29.1 & 3.37 & 501 & 22.9 & 3.61 & 655 & 32.9 & 3.15 \\
        & TimeStream Module & \textbf{372} & \textbf{16.4} & \textbf{3.92} & \textbf{316} & \textbf{11.8} & \textbf{4.08} & \textbf{328} & \textbf{17.5} & \textbf{3.79} \\
        
        \midrule
        
        \multirow{4}{*}{Content Controllability}
        & One-hot Encoding & 491 & 20.9 & 3.71 & 430 & 14.3 & 3.85 & 581 & 25.3 & 3.46 \\
        & Random Encoding & 486 & 21.6 & 3.64 & 446 & 15.1 & 3.79 & 567 & 25.8 & 3.50 \\
        & Learnable Prototypes & 390 & 17.7 & 3.81 & 341 & 13.6 & 3.85 & 396 & 20.8 & 3.55 \\
        & Content-Aware Module & \textbf{372} & \textbf{16.4} & \textbf{3.92} & \textbf{316} & \textbf{11.8} & \textbf{4.08} & \textbf{328} & \textbf{17.5} & \textbf{3.79} \\
        
        \bottomrule
    \end{tabular}
    \label{tab2:ablation1}
\end{table*}
\definecolor{amber(sae/ece)}{rgb}{1.0, 0.49, 0.0}
\begin{figure*}[!htbp]
\centering
\includegraphics[width=1.0\textwidth]{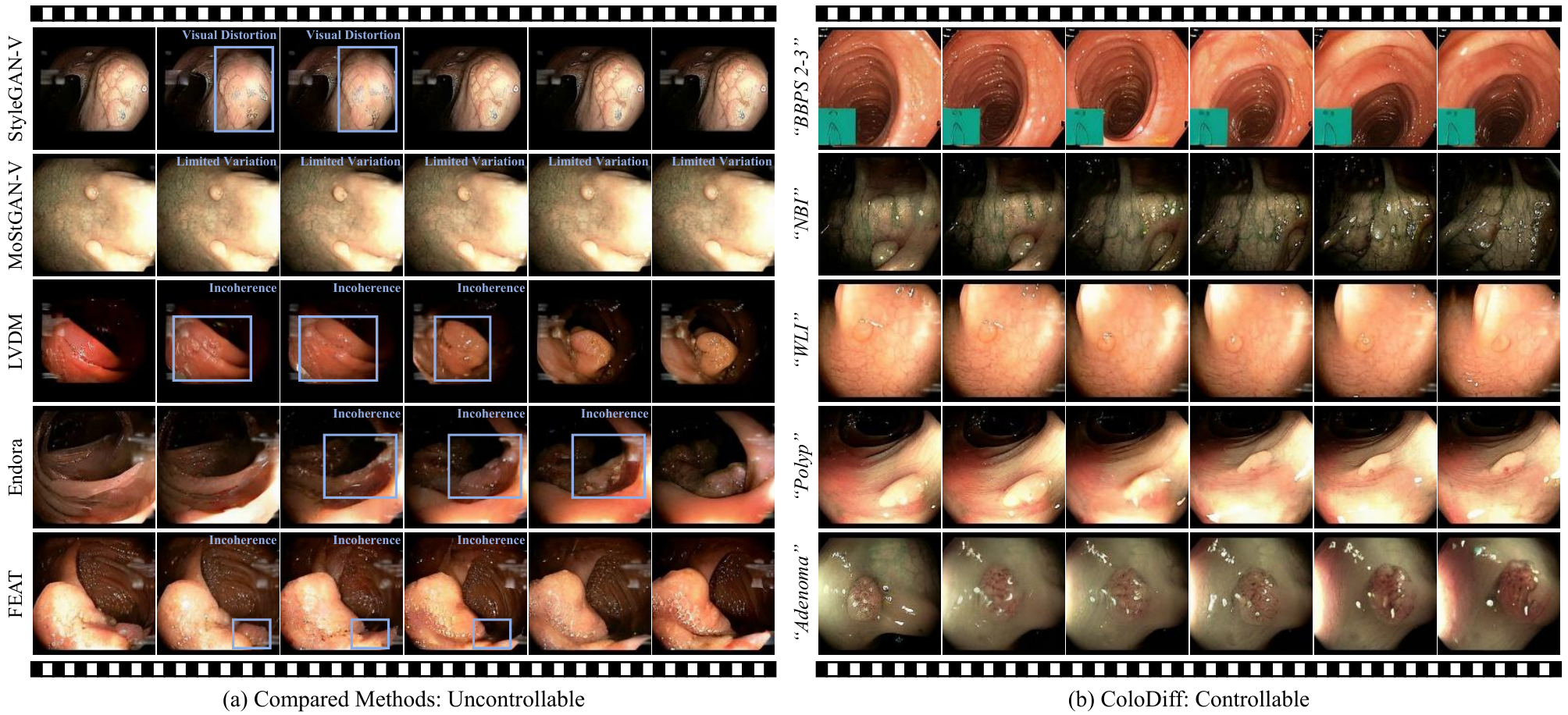}
\caption{Visual comparison of generated videos on Colonoscopic dataset~\cite{colonoscopic}. \bluec{(a)} Videos generated by other compared methods. The \boxbluec{blue} boxes indicate regions that exhibit corresponding issues. Row 1: anatomical and textural visual distortion. Row 2: limited content variation. Rows 3-5: inter-frame incoherence. \bluec{(b)} Videos generated by ColoDiff.}
\label{Fig3_comparison}
\end{figure*}

We reproduce five SOTA generative algorithms for colonoscopy video synthesis~\cite{LVDM,styleganv,mostganv,Endora,FEAT}. Among them, StyleGAN-V~\cite{styleganv} and MoStGAN-V~\cite{mostganv} are GAN-based approaches that achieve high temporal coherence through continuous temporal encoding and motion disentanglement. The other three methods are diffusion-based. LVDM~\cite{LVDM} creatively introduces conditional latent perturbation and unconditional guidance to mitigate error accumulation, thereby producing high-fidelity videos. Endora~\cite{Endora} and FEAT-L~\cite{FEAT} are specially tailored for endoscopic video synthesis, demonstrating superior capability in modeling complex spatio-temporal dynamics and capturing surgical scene representations.

Table~\ref{tab1:comparison} presents the quantitative comparison across four datasets. ColoDiff demonstrates comprehensive superiority over GAN-based approaches~\cite{styleganv,mostganv}, with better FVD, FID, and IS scores. Specifically, ColoDiff achieves a 339 FVD on Colonoscopic, lower than StyleGAN-V~\cite{styleganv} and MoStGAN-V~\cite{mostganv}, while also reducing FID to 12.7 and improving IS to 3.95. Similar trends can be observed on HyperKvasir, SUN-SEG, and the hospital database, confirming that ColoDiff produces colonoscopy videos well aligned with real distribution. In addition, ColoDiff also pushes the boundary of diffusion-based methods~\cite{LVDM,Endora,FEAT}. On the hospital database, it achieves an FVD of 336, substantially outperforming the recently proposed Endora~\cite{Endora} and FEAT-L~\cite{FEAT}, and the results demonstrate ColoDiff’s generalizability to handle complex clinical scenarios. The lower FVD values demonstrate a superior temporal coherence of ColoDiff, while improvements in FID and IS further validate its ability to generate colonoscopy videos with high fidelity and diversity.

As shown in Fig.~\ref{Fig3_comparison}(a), compared methods suffer from visual distortion, limited content variation, and temporal incoherence. The related regions are marked with blue boxes. To be specific, Row 1 displays both anatomical and textural visual distortion, presenting deformed lesion shape and discontinuous vascular topology. Row 2 shows almost no shift in viewpoint or illumination across frames. Rows 3-4 reveal lesions that pop into existence, while Row 5 shows an abrupt lesion disappearance. In contrast, videos in Fig.~\ref{Fig3_comparison}(b) exhibit improved coherence and controllability. This performance stems from the TimeStream module, which decouples temporal dependencies to capture complex intestinal dynamics, and the Content-Aware module, which enhances fine-grained, class-discriminative representation. Additional examples in the \textbf{Supplementary Material} further demonstrate the diversity, coherence, and controllability of ColoDiff-generated videos.

\subsection{Ablation Experiments}
Ablation experiments validate the function of TimeStream and Content-Aware modules from both temporal coherence and content controllability. For temporal coherence, all the compared implementations are combined with Content-Aware module. The baseline is pure Transformer encoder layers. The “TimeStream Module” row means employing interlaced Transformer encoder layers and TimeStream modules. For content controllability, all the compared implementations are combined with TimeStream module. Different category representation approaches are compared, including one-hot encoding, random encoding, and learnable prototypes. These approaches retain the model’s customized content generation ability. The “Content-Aware Module” row means introducing fine-grained guidance besides learnable prototypes. The evaluation metrics include FVD, FID, and IS scores across disease categories to assess video generation quality.

\begin{table}[t]
    \centering
    \setlength{\tabcolsep}{5pt}
    \caption{Ablation Experiment Results for Non-Markovian Sampling Strategy. The Inference Time Column Indicates The Time Required to Generate A 16-frame Video.}
    \begin{tabular}{c c *{3}{c} c c} 
        \toprule
        Time & If & \multicolumn{3}{c}{FID$\downarrow$} & Inference & Frame \\
        \cmidrule(lr){3-5} 
        Steps & Markovian & Colitis & Polyp & Adenoma & Time (s) & Rate \\
        \midrule
        
        $T$=250 & Yes & 16.4 & 11.8 & 17.5 & 13.46 & 1.19 \\
        $T$=100 & No & 16.7 & 12.3 & 17.9 & 5.17 & 3.09 \\
        $T$=50 & No & 17.1 & 12.5 & 18.4 & 2.61 & 6.13 \\
        $T$=10 & No & 17.5 & 13.1 & 18.7 & 0.49 & 32.65 \\
        $T$=5 & No & 19.3 & 14.2 & 21.6 & 0.24 & 66.67 \\
        
        \bottomrule
    \end{tabular}
    \label{tab3:ablation2}
\end{table}
\begin{figure}[t!]
\centering
\includegraphics[width=0.48\textwidth]{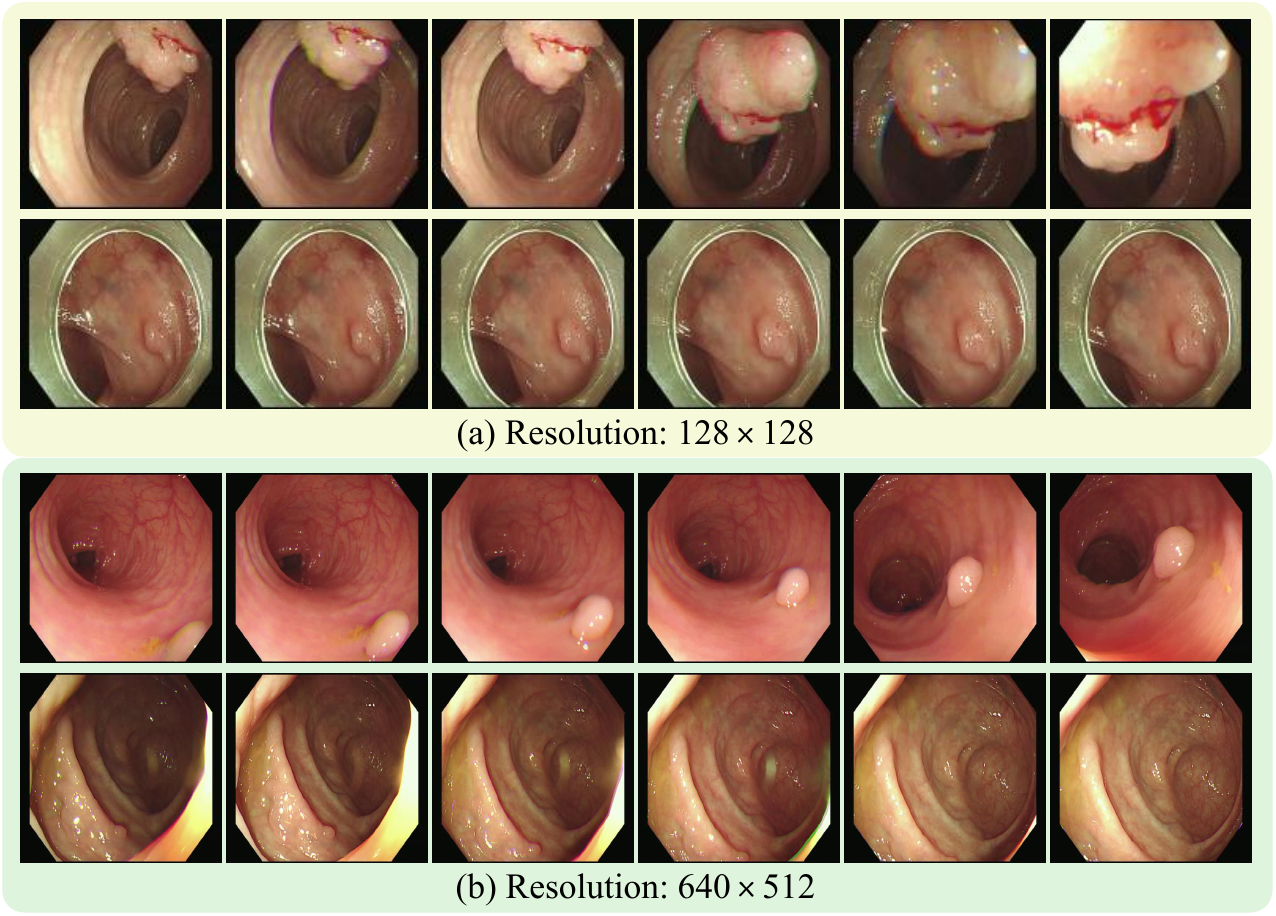}
\caption{Examples of videos generated by ColoDiff under different resolution settings. For uniform display, slight stretching deformation is applied to the original images.}
\label{Fig4_resolution}
\end{figure}
Table~\ref{tab2:ablation1} shows the ablation results. In terms of temporal coherence, TimeStream module achieves remarkable improvements compared with pure Transformer encoder layers. It reduces FVD score by over 20\% across all categories, e.g., from 518 to 372 for colitis. The substantial decrease in FID scores, e.g., from 32.9 to 17.5 for adenoma, also indicates that generated videos achieve better visual fidelity. These results prove that TimeStream module successfully decouples temporal dependencies, capturing motion patterns during endoscope movement and enhancing coherence of colonoscopy videos.

In terms of content controllability, one-hot encoding and random encoding showcase comparable performance, indicating the model’s potential to generate controllable contents based on specific encodings. Prototype learning constructs a learnable feature vector for each category, helping the model reduce FVD for three diseases to below 400. It demonstrates that learnable vectors offer more customized representations compared to fixed encodings. Furthermore, after the noise-injected video embeddings are introduced as fine-grained guidance, the model performs better across all categories and metrics, even achieving a 4.08 IS score for polyp videos. These experimental results substantiate the utility of TimeStream and Content-Aware modules in enhancing inter-frame consistency and content controllability for colonoscopy videos.

In addition, we conduct ablation experiments on the non-Markovian sampling strategy employed by ColoDiff. The number of diffusion time steps are scanned from 250 to 100, 50, 10, and 5. As reported in Table~\ref{tab3:ablation2}, even when the sampling steps are reduced to 50 or 10, the FID scores across all categories do not exhibit an obvious deterioration. For the 10-step setting, ColoDiff averagely generates 32.65 frames per second for 128$\times$128 resolution. Moreover, the Transformer-based design enables ColoDiff’s inherent scalability to fit higher resolution and longer sequence, leading to 26.23 frames per second for 640$\times$512 resolution with half-precision inference, which exceeds the clinical requirements. Fig.~\ref{Fig4_resolution} presents colonoscopy videos generated by ColoDiff at different resolution settings, where the 640$\times$512 resolution yields clearer mucosal and microvascular structures for visual perception. These results indicate that ColoDiff effectively balances inference time and generation quality, pushing diffusion-based video generation toward real-time performance.

\begin{figure}[t!]
\centering
\includegraphics[width=0.48\textwidth]{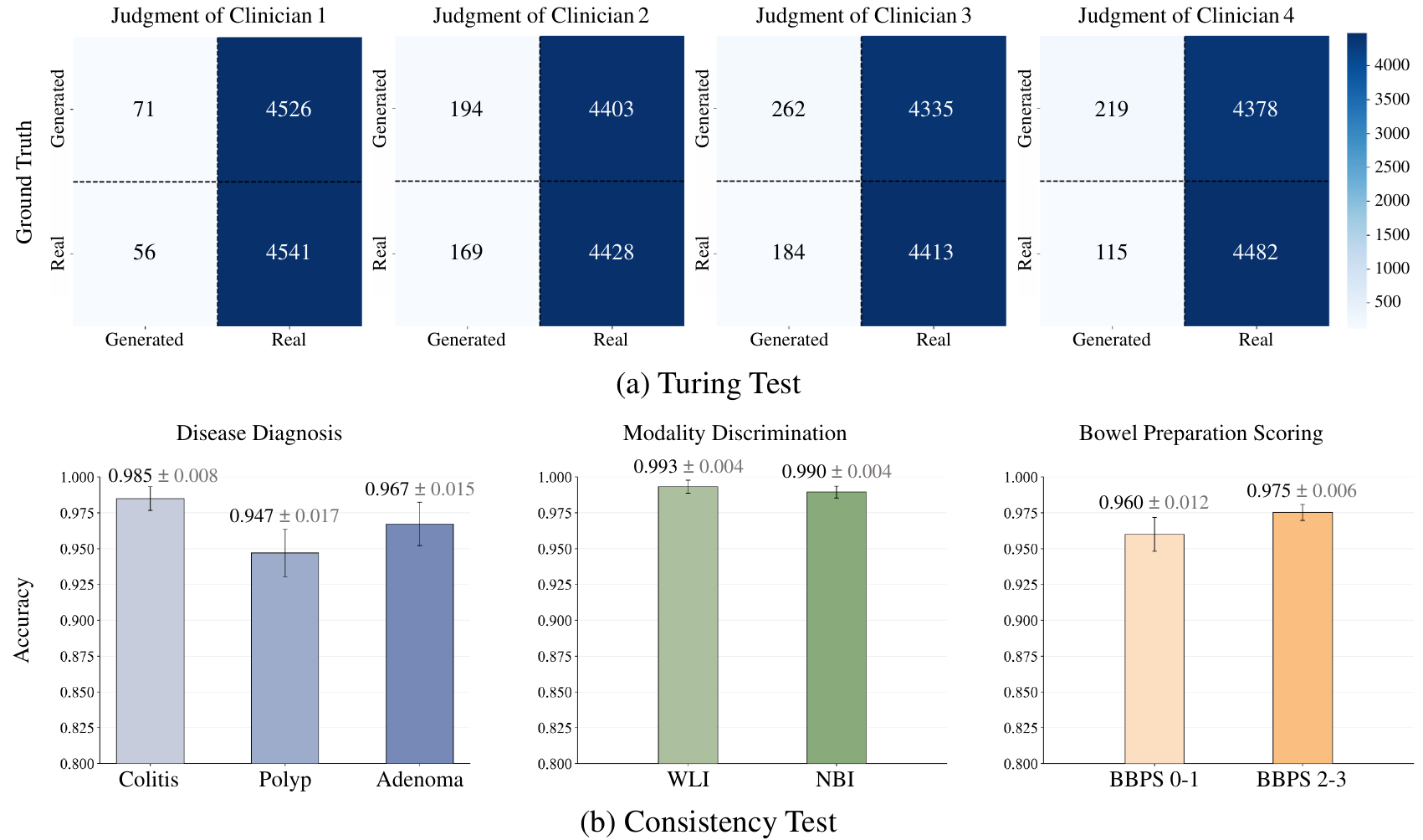}
\caption{The experimental results of Turing test and Consistency test.}
\label{Fig5_clinical}
\end{figure}
\subsection{Clinical Evaluation}
In the clinical domain, professional evaluation and risk assessment are critical to ensuring the reliability of synthetic data. To address this necessity, we invite four clinicians from hospitals to conduct both Turing and Consistency tests. For Turing test, participants are tasked with distinguishing between real and synthetic colonoscopy videos provided at the same resolution. The test set comprises 4,597 real colonoscopy videos and 4,597 synthetic videos generated by ColoDiff, where the ratio is not disclosed to clinicians in advance. For Consistency test, clinicians perform three classification assessments based only on synthetic videos, including disease diagnosis, modality discrimination, and bowel preparation scoring. 1,000 synthetic videos are generated for each category of each task, and the consistency between clinicians’ judgments and ColoDiff’s predefined control conditions is quantified to evaluate content controllability.

Fig.~\ref{Fig5_clinical}(a) and Fig.~\ref{Fig5_clinical}(b) separately display the results of Turing test and Consistency test. In Fig.~\ref{Fig5_clinical}(a), the left two columns show the judgments of two junior clinicians, while the right two columns represent those of two senior clinicians. For the junior clinicians, nearly half of the videos they identify as “generated” are actually real (71 \textit{vs.} \textbf{56}, 194 \textit{vs.} \textbf{169}; \textbf{bold} values indicate real videos). Although the senior clinicians exhibit stronger discrimination ability by identifying 262 and 219 synthetic videos respectively, the strictest clinician (3\textsuperscript{rd} column) still misclassifies over 94.30\% of the synthetic videos as real (4,335 out of 4,597). These results confirm that videos generated by ColoDiff are sufficiently realistic to “pass for real” from the perspective of medical professionals. In Fig.~\ref{Fig5_clinical}(b), the consistency between clinicians’ assessment and ColoDiff’s specified condition further validates the controllability of ColoDiff. Specifically, the average accuracy of four clinicians in disease diagnosis, modality discrimination, and bowel preparation scoring based on synthetic videos is comparable to their performance when evaluating real clinical data. Even for the “polyp” category, where the lowest accuracy was observed in all three tasks, the consistency accuracy still reaches a high value of 0.947 ± 0.017. The experimental results indicate that synthetic videos effectively capture the representative clinical features of different categories, verifying both the fidelity and controllability of ColoDiff’s output.

\begin{table*}[t]
    \centering
    \caption{Quantitative Results for Downstream Classification Tasks (Unit: \%). For Paired Metrics, Models Trained With Real Videos (Left) \textit{vs.} Real + Synthetic Videos (Right) Are Compared. Better Results Are in \textbf{Bold}.}
    \setlength{\tabcolsep}{3pt}
    \small
    \begin{tabular}{
        c
        *{2}{c} 
        *{2}{c} 
        *{2}{c} 
        *{2}{c} 
        *{2}{c} 
        *{2}{c} 
        *{2}{c} 
    }
    \toprule
    \multirow{3}{*}{Metric} & 
    \multicolumn{6}{c}{Disease} & 
    \multicolumn{4}{c}{Modality} & 
    \multicolumn{4}{c}{BBPS} \\
    \cmidrule(lr){2-7} \cmidrule(lr){8-11} \cmidrule(l){12-15}
    & \multicolumn{2}{c}{Colitis} & 
      \multicolumn{2}{c}{Polyp} & 
      \multicolumn{2}{c}{Adenoma} & 
      \multicolumn{2}{c}{WLI} & 
      \multicolumn{2}{c}{NBI} & 
      \multicolumn{2}{c}{0-1} & 
      \multicolumn{2}{c}{2-3} \\
    & \scriptsize Original & \scriptsize +Synthetic & 
      \scriptsize Original & \scriptsize +Synthetic & 
      \scriptsize Original & \scriptsize +Synthetic & 
      \scriptsize Original & \scriptsize +Synthetic & 
      \scriptsize Original & \scriptsize +Synthetic & 
      \scriptsize Original & \scriptsize +Synthetic & 
      \scriptsize Original & \scriptsize +Synthetic \\
    \midrule
    Precision$\uparrow$ & 79.2 & \textbf{85.9} & 85.3 & \textbf{92.8} & 73.5 & \textbf{81.0} & 88.4 & \textbf{93.7} & 93.1 & \textbf{96.8} & 86.7 & \textbf{90.3} & 87.5 & \textbf{93.8} \\
    Recall$\uparrow$    & 82.1 & \textbf{87.4} & 87.7 & \textbf{89.1} & 76.8 & \textbf{83.4} & 91.2 & \textbf{96.3} & 90.6 & \textbf{93.7} & 89.2 & \textbf{92.6} & 85.3 & \textbf{91.8} \\
    F1-score$\uparrow$  & 80.6 & \textbf{86.6} & 86.5 & \textbf{90.9} & 75.1 & \textbf{82.2} & 89.8 & \textbf{95.0} & 91.8 & \textbf{95.2} & 87.9 & \textbf{91.4} & 86.4 & \textbf{92.8} \\
    Accuracy$\uparrow$  & 79.8 & \textbf{85.8} & 85.4 & \textbf{91.7} & 74.3 & \textbf{83.1} & 90.2 & \textbf{94.4} & 91.8 & \textbf{95.1} & 88.5 & \textbf{92.1} & 86.6 & \textbf{92.5} \\
    \bottomrule
    \end{tabular}
    \label{tab4:classification}
\end{table*}

\begin{figure}[t!]
\centering
\includegraphics[width=0.48\textwidth]{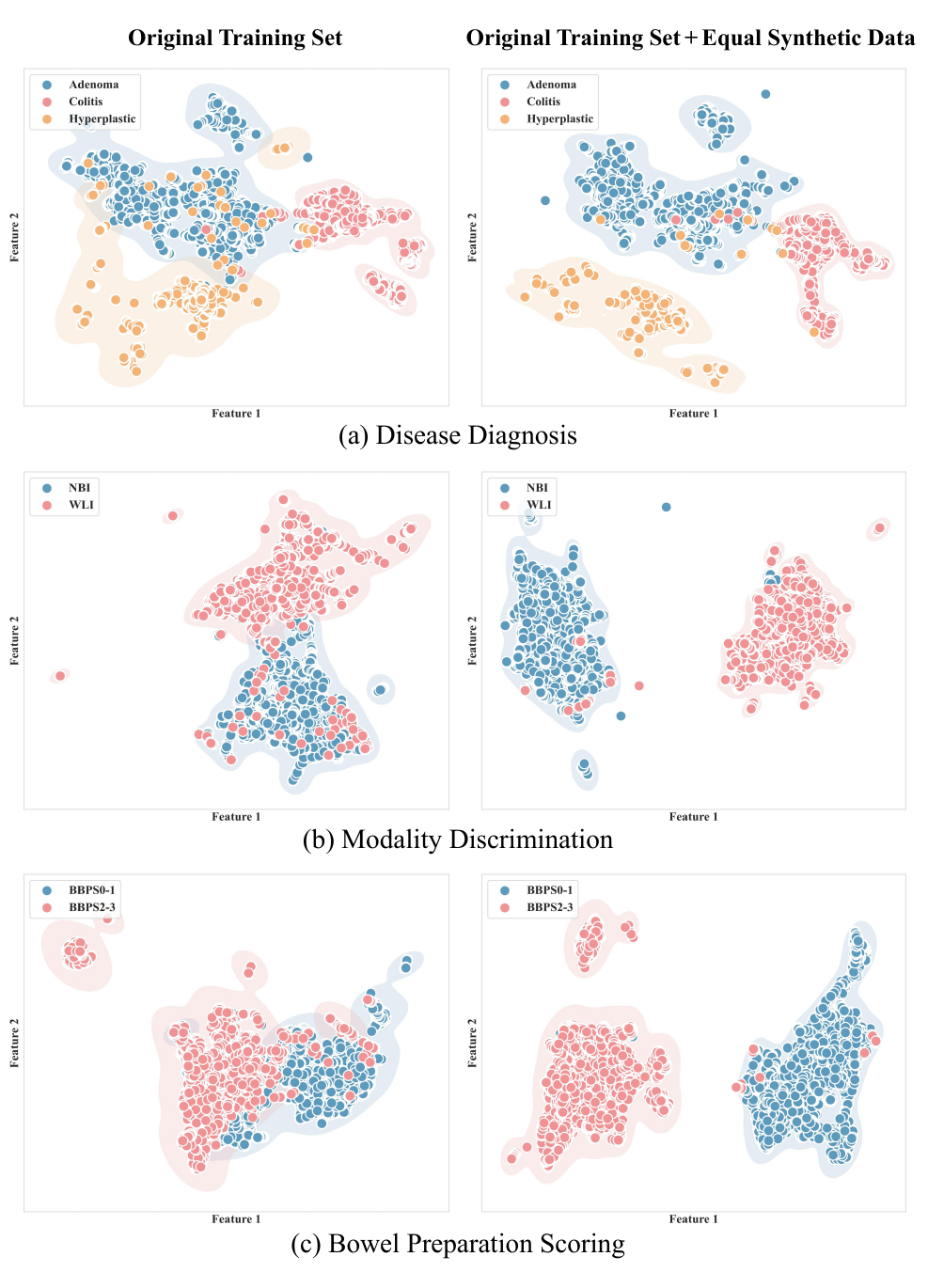}
\caption{UMAP visualization for downstream classification tasks on test sets, with the left column representing models trained with original training set, and the right column representing models trained with an equal number of synthetic videos added.}
\label{Fig6_UMAP}
\end{figure}
\subsection{Downstream Task Experiments}
Benefiting from the Content-Aware module, ColoDiff is capable of generating colonoscopy videos with specified categories, which distinguishes it from other comparative algorithms. Thus, ColoDiff’s synthetic videos can be integrated into the original training data for fully supervised re-training towards various downstream tasks. We conduct classification and segmentation tasks as downstream evaluations to demonstrate the potential applications of ColoDiff. In the experiments, we uniformly adopt 3D ResNet~\cite{3DResNet,ResNet} for classification and SALI~\cite{SALI}, a network dedicated to colonoscopy video dense prediction, for lesion segmentation. SALI uses a 2D architecture to handle video-based segmentation, explicitly modeling inter-frame temporal correlations via short-term alignment and long-term interaction~\cite{SALI}.

\subsubsection{Downstream Classification Tasks}
\begin{figure}[t!]
\centering
\includegraphics[width=0.48\textwidth]{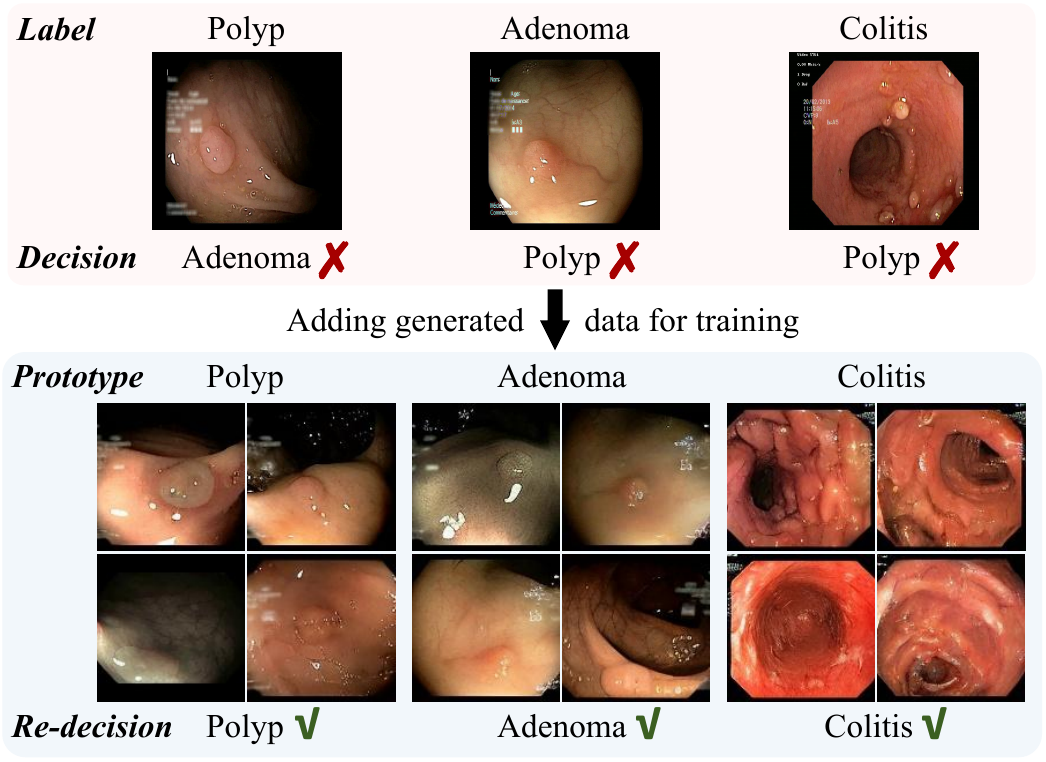}
\caption{Visualized examples of downstream classification performance.}
\label{Fig7_classification}
\end{figure}

Table~\ref{tab4:classification} summarizes the results of three classification tasks, including disease diagnosis, modality discrimination, and bowel preparation scoring (BBPS). The results are reported in pairs: left for models trained only with original videos, and right for those trained with equal synthetic videos added. We find that ColoDiff-synthesized videos consistently improve classification performance for all categories of the three tasks. Especially for disease diagnosis, ColoDiff achieves an average accuracy improvement of 7.1\% (from 79.8\% to 86.9\%). Additionally, we perform the uniform manifold approximation and projection (UMAP)~\cite{UMAP}, a dimension-reduction technique, to map the classifier’s last-layer features into a 2D coordinate system. The left column of Fig.~\ref{Fig6_UMAP} shows that point clusters from different categories have noticeable overlapping areas. In contrast, after incorporating generated videos into training, these clusters are dispersed much better in the right column of Fig.~\ref{Fig6_UMAP}. The visualization results indicate that generated videos enhance feature robustness and inter-class discrimination, owing to their similarity to real data, diffusion-induced randomness, and re-balancing of the training distribution~\cite{NatureBME}.

The upper part of Fig.~\ref{Fig7_classification} shows real cases the classifier initially misclassified. From the constraint of Content-Aware module, ColoDiff can generate videos adhering to specific real-data patterns. With more generated data incorporated into training, the classifier may be guided to make correct re-decisions. Notably, combining category prototypes with diffusion-induced randomness enables ColoDiff to generate non-typical representations, such as adenoma-like polyps, which can challenge the classifier to refine boundary decisions.

\begin{table*}[t]
    \centering
    \caption{Quantitative Results for Downstream Segmentation Tasks (Unit: \%). For Paired Metrics, Models Trained With Real Videos (Left) \textit{vs.} Real + Synthetic Videos (Right) Are Compared. Better Results Are in \textbf{Bold}.}
    \begin{tabular}{ 
        c
        *{8}{c} 
    }
    \toprule
    \multirow{3}{*}{Metric} & 
    \multicolumn{2}{c}{Test-Easy (Seen)} & 
    \multicolumn{2}{c}{Test-Easy (Unseen)} & 
    \multicolumn{2}{c}{Test-Hard (Seen)} & 
    \multicolumn{2}{c}{Test-Hard (Unseen)} \\
    & {\scriptsize Original} & {\scriptsize +Synthetic} 
    & {\scriptsize Original} & {\scriptsize +Synthetic} 
    & {\scriptsize Original} & {\scriptsize +Synthetic} 
    & {\scriptsize Original} & {\scriptsize +Synthetic} \\
    \midrule
    Dice$\uparrow$ & 91.2 & \textbf{93.6} & 81.3 & \textbf{86.5} & 84.5 & \textbf{90.7} & 72.9 & \textbf{84.1} \\
    mIoU$\uparrow$ & 89.3 & \textbf{91.5} & 80.4 & \textbf{85.2} & 88.1 & \textbf{92.6} & 77.4 & \textbf{86.8} \\
    Precision$\uparrow$ & 89.9 & \textbf{93.3} & 80.8 & \textbf{83.1} & 84.9 & \textbf{96.1} & 78.0 & \textbf{86.3} \\
    Recall$\uparrow$ & 92.5 & \textbf{93.9} & 81.8 & \textbf{90.2} & 84.1 & \textbf{85.9} & 68.4 & \textbf{82.0} \\
    \bottomrule
    \end{tabular}
    \label{tab5:segmentation}
\end{table*}

\subsubsection{Downstream Segmentation Tasks}
We also perform segmentation experiments on SUN-SEG dataset~\cite{SUN-SEG}. According to the official distribution: the test set is categorized into \textit{Easy} and \textit{Hard} levels based on difficulty; \textit{Seen} indicates that test images and some training images are from the same video sequence but different frames, while \textit{Unseen} signifies that test images derive entirely from distinct video sequences.

From Table~\ref{tab5:segmentation}, the synthetic-augmented training set improves the average Dice from 82.5\% to 88.7\%, with a 6.2\% increase. Among these results, the performance on \textit{Unseen} subsets is more indicative of the model’s generalization capability. There also exist two notable phenomena. First, for \textit{Easy} data, models exhibit better Dice than mIoU; but for \textit{Hard} data, the situation is reversed. This suggests that edge segmentation for \textit{Hard} images is unsatisfactory; although the model locates lesion areas, it fails to accurately perceive the details. In contrast, after incorporating synthetic data into training, the gap between Dice and mIoU for \textit{Hard} images narrows, indicating an enhancement in the segmentation model's ability to capture lesion edges and textures. Second, segmentation performance on \textit{Unseen} data is much worse than on \textit{Seen} data, as evidenced by 72.9\% Dice for \textit{Unseen} data \textit{vs.} 84.5\% for \textit{Seen} data (\textit{Hard} level). After training with added synthetic data, the Dice score improves to 84.1\% for \textit{Unseen} data \textit{vs.} 90.7\% for \textit{Seen} data, which to some extent reduces this gap and improves the model's robustness to \textit{Unseen} data.

\begin{figure}[t!]
\centering
\includegraphics[width=0.48\textwidth]{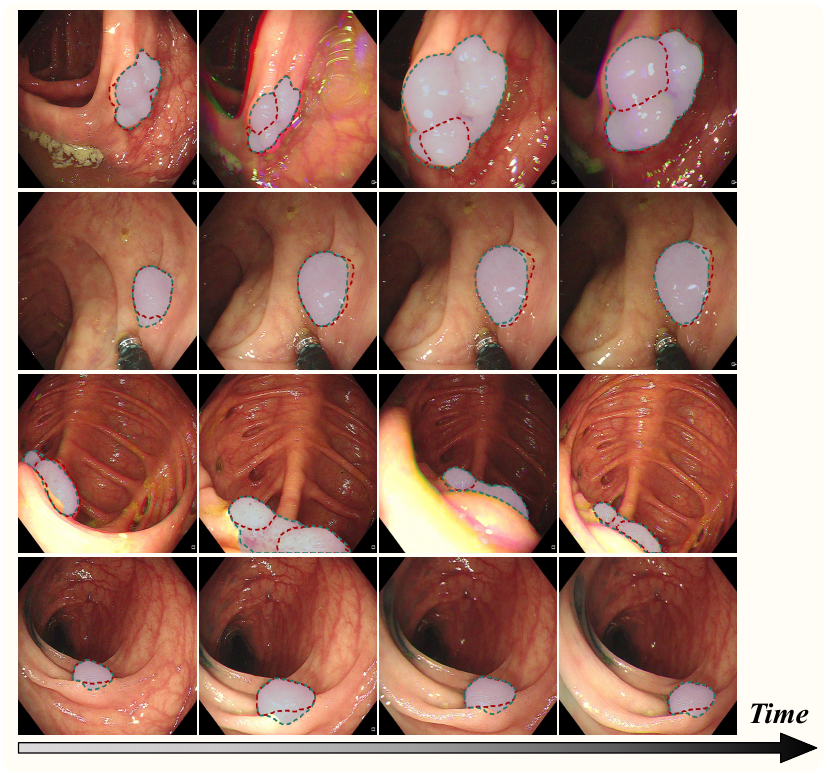}
\caption{Representative examples of downstream segmentation performance. \segbluec{Translucent blue} regions indicate the ground truth, \segredc{red} contours denote results without generated data for training, and \seggreenc{green} contours show results after integrating generated data for training.}
\label{Fig8_segmentation}
\end{figure}

Fig.~\ref{Fig8_segmentation} exhibits sequence-based segmentation examples before and after incorporating synthetic training data. As can be seen, the inclusion of generated data enables the model to segment lesion boundaries more robustly under interference factors like diverse backgrounds, motion artifacts, and specular reflections. The visualization results confirm that generated data facilitates lesion representation learning. Specifically, the randomness of diffusion process, together with the similarity of generated samples to real data, introduces challenging near-boundary cases. By incorporating these cases, the synthetic data re-balances the distribution, regularizes the feature space, and enhances robustness~\cite{NatureBME}, leading to performance gains especially on hard and unseen test conditions.

\section{Discussion}
\label{sec: discussion}
Colonoscopy video generation faces challenges of complex temporal modeling, customized content control, and restricted inference speed. To address these problems, we propose a diffusion-based network with TimeStream and Content-Aware modules, namely, ColoDiff. Experiments show that ColoDiff is capable of generating temporal-coherent and content-controllable videos in real-time scenarios.

\subsection{Towards Temporal-Coherent Colonoscopy Video Generation}
Temporal-coherent colonoscopy video generation is crucial for supplementing high-quality data and assisting diagnosis. As shown in Table~\ref{tab1:comparison}, StyleGAN-V~\cite{styleganv} and MoStGAN-V~\cite{mostganv} show poor FVD scores due to inadequate temporal modeling and training instability. Diffusion-based LVDM~\cite{LVDM} employs 3D U-Net to capture spatio-temporal motion patterns. Constrained by CNN’s locality bias, it does not achieve long-range dependency in temporal dimension, leading to inter-frame inconsistency (Row 3, Fig.~\ref{Fig3_comparison}(a)). Although other diffusion-based methods employ Transformer architectures~\cite{Endora,FEAT}, the proposed strategies like fully cross-frame interaction excessively use the model’s scaling ability~\cite{ControlVideo}, leading to the “forgetting” problem like abrupt lesion appearance or disappearance (Rows 4-5, Fig.~\ref{Fig3_comparison}(a)). 

In contrast, ColoDiff achieves the best FVD scores across all datasets, e.g., 294 on SUN-SEG dataset, which is 17.4\% lower than the best compared method~\cite{FEAT}. The results indicate that generated videos maintain high temporal coherence even in the presence of irregular intestinal structures. This advantage is largely attributed to our TimeStream module, which treats spatially aligned patches as sequential input tokens. By leveraging the prior knowledge that the same anatomical structure generally occupies consistent or adjacent spatial locations across consecutive frames, the TimeStream module enables accurate modeling of irregular morphology and complex motion. Moreover, by equipping 2D architectures with 3D contextual reasoning capabilities, it empowers ColoDiff to work efficiently without increasing model scale or computational cost.

\subsection{Towards Content-Controllable Colonoscopy Video Generation}
Content-controllable generation facilitates customized data synthesis, thus benefiting various downstream tasks. Based on Table~\ref{tab2:ablation1}, prototype learning outperforms one-hot and random encoding, indicating the superiority of learnable embeddings for multi-pattern conditioning. This is because prototype learning assigns a unique but adjustable vector to each category. As ColoDiff progressively learns to fit noise patterns specific to different categories, the vectors become class-discriminative. Meanwhile, standard diffusion models only rely on the time-step index $t$ to perceive noise levels, yielding coarse conditioning without spatio-temporal information. To overcome this, noise-injected data embedding is further employed as a fine-grained regulation with intra-frame spatial information. Integrating prototypes with noise-injected embeddings, the Content-Aware module refines ColoDiff’s multi-layer features, pushing the performance boundary of polyp-IS to 4.08. 

According to Tables~\ref{tab4:classification} and~\ref{tab5:segmentation}, the randomness of diffusion process, combined with the similarity of generated samples to real data, yields feature variations while preserving anatomical fidelity. Incorporating such data into training successfully improves downstream task performance, confirming that ColoDiff produces diverse yet controllable colonoscopy videos. This can be further supported by the UMAP visualization in Fig.~\ref{Fig6_UMAP}, where feature clusters from different categories become more separated after incorporating synthetic data. The application of customized prototypes to provide specific feature representations for different data patterns can be adopted by other generative models. The use of noise-injected embedding as supplementary information to time-step index $t$, thereby enhancing the model controllability, can also be applied to other diffusion-based models.

\subsection{Future Work}
In our future work, improving the controllability of generative models remains a key focus. Our ambition is to achieve concurrent regulation over multiple variables, including imaging modality, disease type, etc. Additionally, we aim to compile comprehensive video-text datasets to facilitate more precise control over colonoscopy video generation through multi-modal alignment.

\section{Conclusion}
\label{sec: conclusion}
This paper proposes ColoDiff, a diffusion-driven approach with TimeStream and Content-Aware modules for colonoscopy video generation. ColoDiff successfully decouples temporal dependency across frames and precisely controls clinical attributes within frames. The non-Markovian sampling strategy further enables real-time synthesis. Comprehensive experiments on multiple benchmarks and downstream tasks confirm ColoDiff’s ability to generate realistic, coherent, and controllable colonoscopy videos. Incorporating synthetic data into training improves disease diagnosis accuracy by 7.1\% and lesion segmentation Dice by 6.2\%. Our research presents an exploration in leveraging generated videos for colonoscopy analysis, paving the way for future integration of synthetic data into clinical practice.

\bibliographystyle{IEEEtran}
\bibliography{references}

\end{document}